# S-Shaped vs. V-Shaped Transfer Functions for Ant Lion Optimization Algorithm in Feature Selection Problem

Majdi Mafarja
Department of Computer Science,
Birzeit University
P.O. Box 14
Palestine
mmafarja@birzeit.edu

Derar Eleyan
Department of Applied Computing,
Palestine Technical University
P.O. Box 7,
Department of Computer Science,
Birzeit University
Palestine

Salwani Abdullah
Data Mining and Optimisation
Research, Group (DMO)
43600 UKM
Malaysia
salwani@ukm.edu.my

Seyedali Mirjalili
School of Information and
Communication Technology
Griffith University
QLD 4111
Australia
seyedali.mirjalili@griffith.edu.au

## ABSTRACT

Feature selection is an important preprocessing step for classification problems. It deals with selecting near optimal features in the original dataset. Feature selection is an NP-hard problem, so meta-heuristics can be more efficient than exact methods. In this work, Ant Lion Optimizer (ALO), which is a recent metaheuristic algorithm, is employed as a wrapper feature selection method. Six variants of ALO are proposed where each employ a transfer function to map a continuous search space to a discrete search space. The performance of the proposed approaches is tested on eighteen UCI datasets and compared to a number of existing approaches in the literature: Particle Swarm Optimization, Gravitational Search Algorithm and two existing ALO-based approaches. Computational experiments show that the proposed approaches efficiently explore the feature space and select the most informative features, which help to improve the classification accuracy.

## CCS CONCEPTS

- **Mathematical optimization** →*Discrete optimization;*
- **Artificial intelligence -> Search methodologies** → *Heuristic function construction;*
- **Machine learning** → **Machine learning algorithms** → *Feature Selection*

## 1. INTRODUCTION

Feature Selection (FS) plays a vital role in machine learning since it aims to reduce the data size by eliminating the irrelevant/redundant features from the original datasets [1]. The use of FS algorithms in conjunction with a classification algorithm improves the classification accuracy and/or reducing the processing time [2]. FS methods can be classified based on two main criteria; searching the feature space for the near optimal feature subset and the evaluation of the selected subsets. Evaluating the selected subsets can be classified into two different approaches; Filter and Wrapper. A filter approach usually evaluates the subset depending on the data itself, whereas a wrapper approach uses an external learning algorithm (mostly machine learning technique) to evaluate the selected features.

Searching a feature space to find the best combination of features is an NP-hard problem [1]. Therefore, using brute-force search techniques are impractical with FS problems especially for the medium and large-scale datasets. As an alternative, heuristic methods can be used to find the near-optimal subset faster than brute-force methods [3]. Evolutionary Computation (EC) are population-based meta-heuristic algorithms with a global search capability [4]. Most of such methods are nature-inspired and mimic the behavior (social and biological) of animals, insects or birds like whales, bees, ants, antlions, bats, etc.[5]. In EC, a population of individuals (solutions) interact to obtain the optimal solution [4]. EC have been widely employed to tackle FS problems in the literature [6]. For instance, there are feature selection methods based on Genetic Algorithm (GA) [7, 8], Particle Swarm Optimization (PSO) [9], Ant Colony Optimization (ACO) [10, 11], Differential Evolution (DE) [12], and Artificial Bee Colony (ABC)

[13] For a comprehensive list of works in this field, readers can refer to [14-18].

Recently, new EC algorithms are proposed and have shown good performances when dealing with the FS problem. For instance, a wrapper feature selection model that was based on Ant Lion Optimization Algorithm (ALO) [19] proposed in [20] and [6], and Grey Wolf Optimizer (GWO) [21] has been successfully employed for solving feature selection problems in [22, 23].

ALO is a recent EC algorithm that mimics the behavior of antlions in hunting preys. An ALO-based wrapper FS method has been proposed by Zawbaa *et al.* in [20]. The proposed approach was compared with PSO- and GA-based algorithms and showed competitive performance. Another work in the literature is a chaotic-based ALO algorithm [24], in which a set of chaotic functions were used to control the balance between exploration and exploitation in the original algorithm. Later on, a binary version of the ALO algorithm was proposed by Emary *et al.* in [6]. In all the aforementioned works, two transfer functions (proposed by Kennedy and Eberhart [18] and Rashedi *et al.* [19],) were used to convert the continuous ALO to a binary version with the eventual goal of solving feature selection problems.

According to Mirjalili and Lewis in [25], a transfer function is an important part in the binary versions of the metaheuristics. It significantly impacts the local optima avoidance and the balance between exploration and exploitation. In this paper, six transfer functions, proposed in [25], are applied to the ALO algorithm. A total of three s-shaped and three v-shaped are employed as the first attempt in the literature to find a suitable transfer function for the ALO algorithm. A wrapper model that uses K-Nearest Neighborhood (KNN) classifier is adopted as an evaluation criterion in this work. The results are compared with the two variants of basic ALO algorithm, PSO [26], and GSA [27] for verification.

The rest of this paper is organized as follows: The basic ALO algorithm is presented in Section 2. Section 3 presents the details of the proposed approaches. In Section 4, the experimental results are presented and analyzed. Finally, conclusions and future work are given in Section 5,

## 2. BINARY ALO ALGORITHM

ALO is a recent EC algorithm proposed by Mirjalili [19]. ALO algorithm mimics the interaction between the antlion insects and ants in the hunting process. In nature, an antlion digs a trap with a cone shape. The size of the trap is directly proportional to the hunger level of an antlion. Then, the antlion hides underneath the bottom of the trap waiting for an ant, which moves randomly around the trap, to fall down. Once the antlion realizes that there is an ant in the trap, it catches it. Based on this brief description of the antlion hunting process, the following items can be formulated as a set of conditions for the overall process [19]:

- The ants move in a random walk in the search space. The random walk of the ants at each iteration of the algorithm is simulated as in Eq. (1)

$$X(t) = [0, cumsum(2r(t1) - 1), cumsum(2r(t1) - 1), \ldots, cumsum(2r(tn) - 1) ] \quad (1)$$

where cumsum represents the cumulative sum, n is the max iteration, t is the iteration and $r(t)$ is a stochastic function that takes value (1) if a random number is less than 0.5 and 0 otherwise.

- The traps of antlions affect the moves of the ants in the search space. Eq. 2 and Eq. 3 model this assumption:

$$c_i^t = Antlion_j^t + c^t \quad (2)$$

$$d_i^t = Antlion_j^t + d^t \quad (3)$$

where $c^t$ and $d^t$ are two vectors that contain the minimum and maximum of all variables in *t-th* iteration, $c_i^t$ and $d_i^t$ are the minimum and maximum *i-th* ant and $Antlion_j^t$ represents the position of the *j-th* antlion at the *t-th* iteration.

- The largest trap belongs to the fittest antlion. Thus, catching an ant by an antlion is proportional to the fitness of that antlion (i.e., the antlion with the higher fitness has a higher chance to catch an ant). To model this assumption, a selection mechanism based on a roulette wheel operator is used.
- To model sliding ants towards antlions, the radius of random walks of the ant is decreased adaptively using Eq. (4) and Eq. (5).

$$c^t = c^t/I \quad (4)$$

$$d^t = d^t/I \quad (5)$$

where $I$ is a ration that controls the exploration/exploitation rate in ALO algorithm by limiting the random walk range of the ants and preys. The parameter $I$ in the above equations is defined in Eq. 6.

$$I = 10^w \frac{t}{T} \quad (6)$$

where $t$ is the current iteration, $T$ is the max iteration, and $w$ is a constant that can adjust the accuracy level of the exploitation. $w$ is defined based on the current iteration ($w = 2$ when $t > 0.1T$, $w = 3$ when $t > 0.5T$, $w = 4$ when $t > 0.75T$, $w = 5$ when $t > 0.9T$, and $w = 6$ when $t > 0.95T$).

- If an ant is caught and pulled under the sand by the antlion, then it becomes fitter than its corresponding antlion. Then, the antlion updates its position to the latest caught prey and builds a trap to improve its chance of catching another prey after each hunt. Eq. (7) models this process:

$$Antlion_j^t = Ant_i^t \text{ if } f(Ant_i^t) \text{ is better than } f(Antlion_j^t) \quad (7)$$

where $t$ represents the current iteration, $Antlion_j^t$, $Ant_i^t$ represent the position of the *j-th* antlion and the *i-th* ant at the *t-th* iteration.

The antlion with the higher fitness in each iteration is considered as the elite. The elite antlion ($E$) and the selected antlion by using the selection mechanism ($S$) guide the random walk of an ant ($RW_1$ and $RW_2$ respectively). Since these two solutions are continuous, they must be converted to a binary version to suit the feature selection problem. The conversion is performed by applying different transfer functions that belong to two families (s-shaped and v-shaped) [25]. The transfer function defines the probability of updating the binary solution's elements from 0 to 1 and vice versa. In S-shaped functions, the solution is updated based on Eq. (8)

$$x_i^d(t+1) = \begin{cases} 1 & \text{If } rand < Sfunction\left(x_i^d(t+1)\right) \\ 0 & \text{If } rand \geq Sfunction\left(x_i^d(t+1)\right) \end{cases} \quad (8)$$

where $x_i^k(t+1)$ is the *i-th* element in the solution $x$ at dimension $d$ calculated as $RW_1 - E$ or $RW_2 - S$, $rand$ is a random number drawn from uniform distribution $\in [0,1]$.

on the position updating using a v-shaped transfer function should be done by Eq. (9)

$$X_{t+1} = \begin{cases} \neg X_t, & rand < Vfunction(x_{t+1}) \\ X_t, & rand \geq Vfunction(x_{t+1}) \end{cases} \quad (9)$$

In this work, we are interested to study the impacts of different transfer functions on the performance of ALO. As such, six transfer functions proposed by Mirjalili and Lewis [25] are employed and replaced by the two transfer functions used in [6]. Table 1 shows the mathematical formulation of the transfer functions used in this paper and Fig. 2 shows these two families of transfer functions.

**Table 1. V-shaped and S-shaped transfer functions [25]**

| | | Mathematical formulation |
|---|---|---|
| V-shaped transfer functions | | |
| 1 | ALO-V1 | $T(x) = \left\|\text{erf}\left(\frac{\sqrt{\pi}}{2}x\right)\right\| = \left\|\frac{\sqrt{2}}{\pi}\int_0^{\frac{\sqrt{\pi}}{2}x} e^{-t^2}\,dt\right\|$ |
| 2 | ALO-V2 | $T(x) = \left\|\frac{x}{\sqrt{1+x^2}}\right\|$ |
| 3 | ALO-V3 | $T(x) = \left\|\frac{2}{\pi}\arctan\left(\frac{\pi}{2}x\right)\right\|$ |
| S-shaped transfer functions | | |
| 4 | ALO-S1 | $T(x) = \frac{1}{1+e^{-2x}}$ |
| 5 | ALO-S2 | $T(x) = \frac{1}{1+e^{-x}}$ |
| 6 | ALO-S3 | $T(x) = \frac{1}{1+e^{\frac{-x}{3}}}$ |

## 3. BINARY ALO FOR FEATURE SELECTION

In FS problems, a solution is represented as an $N$-sized binary vector, where $N$ is the total number of features in a dataset. The complexity of generating all possible feature combinations would be $2^N$ where a brute-force search becomes impractical. Meta-heuristics are more reliable techniques for such problems. ALO is one of the metaheuristic that show a good performance in searching the feature space for the best feature subset. Each feature subset is evaluated according to two criteria; number of selected features in addition to the classification accuracy obtained when using those features. The fitness function that takes into consideration those two criteria is modeled in Eq. (10).

$$Fitness = \alpha \gamma_R(D) + \beta \frac{R}{N}, \quad (10)$$

where $\gamma_R(D)$ represents the classification error rate of a given classier (the K-Nearest Neighbor (KNN) classifier [28] is used here). $|R|$ is the number of selected features, $|N|$ is the total number of features in the dataset, and $\alpha \in [1,0]$, $\beta = (1-\alpha)$ are two parameters corresponding to the importance of classification quality and subset length as per the recommendations in [6]. Algorithm 1 shows the pseudocode of the proposed approach.

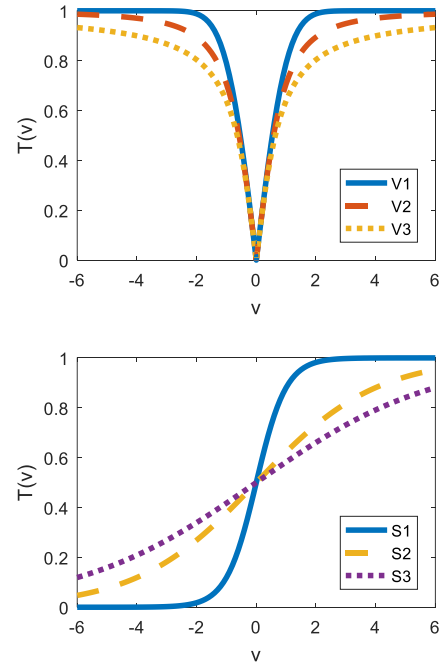

**Figure 2. Transfer functions families (top) v-shaped and (bottom) s-shaped** [25]

**Algorithm 1:** The pseudocode of the binary ant lion optimizer

Initialize *n* ants (ant population) randomly
Initialize *n* antlions (antlion population) randomly
Evaluate each ant and antlion in the two populations
Mark the fittest antlion as Elite
For 1 to max number of iterations
    Calculate ($I$), the radius of random walk of the ant using Eqs. (4), (5) and (6).
    For each $ant_j$ in the population do
- Select an antlion using RWS (AntRW)
- Apply random walk around AntRW to produce RA
- Apply random walk around Elite to produce RE
- Convert RA and RE from continuous to binary using transfer functions to produce RW1 and RW2
- Update the position of $ant_j$ by performing crossover between $RW_1$ and $RW_2$

    End
    Update the fitness of all ants
    Update the position and fitness of antlion based on the position of its corresponding ant if it becomes fitter.
    Update the Elite's position if any antlion becomes fitter than it.
End
Output the Elite antlion and its fitness.

The ALO algorithm starts by generating two random populations for the antlions and ants. All individuals in the populations are evaluated and the best antlion is marked as the elite. Until a stopping criterion satisfied, each ant in the population updates its position with respect to either the elite solution or the solution selected by the roulette wheel. Two continuous solutions (RW1 and RW2) are generated for each ant. Hence, a transfer function is used to convert them to a binary format by defining a probability of updating the binary solution's elements from 0 to 1 and vice versa. Finally, the position of ant is updated by performing crossover between RW1 and RW2. If the fitness of the ant becomes better than that for the corresponding antlion, then the antlion's position is updated based on it.

In this study, six transfer functionsare embedded in the ALO. The three approaches that use s-shaped transfer functions are named ALO-S1, ALO-S2, and ALO-S3, while the three approaches named ALO-V1, ALO-V2, and ALO-V3 use v-shaped transfer functions.

## 4. EXPERIMENTAL RESULTS AND DISCUSSION

In this paper, a wrapper feature selection method based on BLAO is proposed. KNN is chosen as the classifier, and a Euclidean distance matrix is employed to evaluate the algorithm. Each algorithm is run 20 times with a random seed on an Intel Core i5 machine, 2.2 GHz CPU and 4 GB of RAM. Eighteen well-known benchmark datasets from the UCI data repository [29] are used to assess the performance of the proposed approaches. The details of the datasets including number of attributes and instances are shown in Table 1.

**Table 1. The datasets used in the experiments**

| Dataset | No. of Attributes | No. of Objects |
|---|---|---|
| Breastcancer | 9 | 699 |
| BreastEW | 30 | 569 |
| CongressEW | 16 | 435 |
| Exactly | 13 | 1000 |
| Exactly2 | 13 | 1000 |
| HeartEW | 13 | 270 |
| IonosphereEW | 34 | 351 |
| KrvskpEW | 36 | 3196 |
| Lymphography | 18 | 148 |
| M-of-n | 13 | 1000 |
| PenglungEW | 325 | 73 |
| SonarEW | 60 | 208 |
| SpectEW | 22 | 267 |
| Tic-tac-toe | 9 | 958 |
| Vote | 16 | 300 |
| WaveformEW | 40 | 5000 |
| WineEW | 13 | 178 |
| Zoo | 16 | 101 |

The proposed methods are compared PSO [26], GSA [27], and two basic ALO algorithms (coded as bALO1 [6] and bALO2 [6]). Note that, bALO1 uses an s-shaped transfer function and bALO2 uses a v-shaped transfer function. The initial parameters and other experimental setup are presented in Table 2.

**Table 2. Parameter setting for experiments**

| Parameter | Value |
|---|---|
| K for cross validation | 10 |
| Number of runs | 20 |
| Population size | 8 |
| Number of iterations | 70 |
| Problem dimension | No. of features in the dataset |
| Inertia weight for PSO [6] | 0.1 |
| Individual-best acceleration factor of PSO [6] | 0.1 |
| $\alpha$ parameter in the fitness function [6] | 0.99 |
| $\beta$ parameter in the fitness function [6] | 0.01 |
| $G_0$ for GSA [30] | 100 |
| $\alpha$ For GSA [30] | 20 |

All algorithms are compared based on three criteria: classification accuracy, average number of selected features and the average computational time.

Inspecting Table 3, its evident that the approaches using v-shaped transfer functions are better than those equipped with s-shaped transfer functions in terms of classification accuracy (denoted in bold). This enhancement in performance may be interpreted by the abrupt switching between 0 and 1 in case of using v-shaped function that emphasize the explorative behavior of an algorithm. In addition, it may be seen that ALO-based approaches are better than GSA and PSO over all datasets. It is worth mentioning that ALO-V3 approach provides the best results on 11 datasets and outperforms GSA and PSO on all datasets, while the other proposed approaches in this work show competitive results on the majority of case studies.

Table 3 also shows that ALO-based approaches perform better than PSO and GSA approaches. The reason is that ALO algorithm contains only one parameter (I) that controls the balance between exploration and exploitation. This parameter enforces a high level of exploration at the beginning of the searching process while more exploitation is imposed at the end of the optimization process. This is a bonus in avoiding local solutions. Another reason of ALO's superiority is its ability to search the feature space for the most informative features. Each ant is updated based on the best solution so far (a mechanism to highlight exploitation), and based on a solution that selected by the roulette wheel (an operator to support exploration). This gives more chances to the weak solutions to be involved in the searching process with the hope to find promising areas in the search space.

Table 5 reports the computational time to obtain near optimal feature subset. Note that we re-implemented the PSO, GSA, bALO1 and bALO2 algorithms in Matlab and used the same parameter settings on all datasets to provide a fair comparison. It can be seen that ALO-V2 approach has the lowest computation time over 14 datasets. Overall ALO based approaches are better than GSA and PSO approaches in terms of computational time. This again proves the superiority of ALO algorithm on the FS problem.

Table 4 includes the average number of selected features obtained from different algorithms when solving the case studies. It is observed that ALO-V3 shows better performance than other approaches on the majority of the datasets in terms of the number of selected attributes. It obtains the minimal number of selected attributes in eight datasets, while there is no other approach that could outperform other approaches in more than four datasets. The same observation can be made when analyzing the average selection ratio and classification accuracy for the same dataset; ALO-V3 has the best performance in both of them. The remarkable increase in the ALO's results when employing different transfer functions indicates the important role of the transfer function in improving the performance of the ALO algorithm.

**Table 3. Average classification accuracy obtained from different optimizers**

| Dataset | ALO-S1 | ALO-S2 | ALO-S3 | ALO-V1 | ALO-V2 | ALO-V3 | bALO1 | bALO2 | GSA | PSO |
|---|---|---|---|---|---|---|---|---|---|---|
| Breastcancer | 0.950 | 0.951 | 0.949 | 0.972 | 0.963 | **0.974** | 0.951 | 0.973 | 0.942 | 0.957 |
| BreastEW | 0.936 | 0.938 | 0.936 | **0.975** | 0.954 | 0.974 | 0.934 | 0.974 | 0.940 | 0.945 |
| Exactly | 0.646 | 0.640 | 0.637 | 0.982 | 0.710 | 0.965 | 0.645 | **0.985** | 0.697 | 0.683 |
| Exactly2 | 0.712 | 0.702 | 0.698 | **0.762** | 0.734 | **0.762** | 0.702 | **0.762** | 0.715 | 0.721 |
| HeartEW | 0.742 | 0.739 | 0.732 | 0.838 | 0.793 | 0.838 | 0.741 | **0.839** | 0.761 | 0.790 |
| Lymphography | 0.700 | 0.701 | 0.732 | 0.916 | 0.730 | **0.917** | 0.715 | 0.913 | 0.672 | 0.698 |
| M-of-n | 0.707 | 0.690 | 0.735 | 0.964 | 0.864 | 0.967 | 0.711 | **0.969** | 0.714 | 0.818 |
| PenglungEW | 0.745 | 0.739 | 0.750 | **0.827** | 0.730 | **0.827** | 0.745 | 0.826 | 0.630 | 0.666 |
| SonarEW | 0.713 | 0.732 | 0.732 | 0.852 | 0.731 | 0.845 | 0.734 | **0.856** | 0.760 | 0.775 |
| SpectEW | 0.800 | 0.805 | 0.790 | **0.900** | 0.806 | 0.899 | 0.796 | 0.898 | 0.754 | 0.747 |
| CongressEW | 0.909 | 0.911 | 0.902 | 0.980 | 0.940 | **0.981** | 0.903 | 0.980 | 0.913 | 0.931 |
| IonosphereEW | 0.832 | 0.840 | 0.839 | 0.900 | 0.852 | **0.904** | 0.830 | 0.896 | 0.837 | 0.843 |
| KrvskpEW | 0.758 | 0.758 | 0.803 | 0.974 | 0.911 | 0.973 | 0.786 | **0.975** | 0.783 | 0.849 |
| Tic-tac-toe | 0.684 | 0.666 | 0.655 | 0.779 | 0.766 | **0.783** | 0.654 | 0.778 | 0.666 | 0.706 |
| Vote | 0.884 | 0.886 | 0.880 | 0.970 | 0.920 | **0.972** | 0.901 | 0.969 | 0.904 | 0.919 |
| WaveformEW | 0.691 | 0.704 | 0.695 | 0.793 | 0.763 | **0.797** | 0.676 | 0.791 | 0.693 | 0.743 |
| WineEW | 0.857 | 0.886 | 0.877 | 0.971 | 0.921 | **0.972** | 0.890 | 0.969 | 0.864 | 0.930 |
| Zoo | 0.835 | 0.862 | 0.856 | 0.976 | 0.902 | **0.980** | 0.888 | **0.980** | 0.752 | 0.781 |

**Table 4. Average number of selected features obtained from different optimizers**

| Dataset | ALO-S1 | ALO-S2 | ALO-S3 | ALO-V1 | ALO-V2 | ALO-V3 | bALO1 | bALO2 | GSA | PSO |
|---|---|---|---|---|---|---|---|---|---|---|
| Breastcancer | 5.25 | 4.55 | 4.45 | **4.35** | 5.6 | 4.7 | 4.65 | 4.4 | 6.85 | 7 |
| BreastEW | 16.5 | 17.05 | 14.6 | 14.5 | 18.5 | **13.85** | 16.1 | 14.9 | 17.2 | 15.05 |
| Exactly | 7.95 | 7.75 | 7.75 | 6 | 11.4 | **5.75** | 7.7 | 6 | 7.45 | **5.75** |
| Exactly2 | 2.05 | 2.3 | 3.05 | 1.8 | 11 | **1.5** | 2.5 | **1.5** | 2.7 | 2.4 |
| HeartEW | 8.55 | 8 | 8.1 | **7.25** | 8.8 | 8.6 | 8.9 | 7.45 | 8.45 | 7.7 |
| Lymphography | 8.15 | 8.7 | 8.2 | **7.15** | 8.45 | 7.35 | 8.4 | 7.8 | 9.65 | 9.2 |
| M-of-n | 8.75 | 8.2 | 8.4 | 6.1 | 12.05 | **6** | 8.5 | **6** | 7.5 | **6** |
| PenglungEW | 140.1 | 153.75 | 153.3 | 135 | 156.2 | 133.1 | 146.9 | **130.15** | 153 | 149.4 |
| SonarEW | **26.1** | 28 | 27.6 | 26.8 | 28.6 | 26.6 | 28.1 | 26.5 | 29.7 | 27 |
| SpectEW | 8.25 | 8.9 | 9.25 | **6.95** | 10.35 | 7.65 | 9.1 | 7.25 | 10.85 | 9.65 |
| CongressEW | 7.4 | 7.9 | 8.35 | 7.15 | 9.1 | **6.65** | 7.55 | 6.95 | 7.55 | **6.65** |
| IonosphereEW | 12.55 | 13.15 | 14.4 | 11.35 | 19.55 | 11.75 | 13.5 | 11.6 | 12.9 | **10.25** |
| KrvskpEW | 19.55 | 18.25 | 18.75 | 16.25 | 27.85 | **16.15** | 20.15 | 16.9 | 19.85 | 17.1 |
| Tic-tac-toe | 4.9 | 4.95 | 4.9 | 5 | 8.05 | 5 | **4.85** | 5 | 5.9 | 6 |
| Vote | 7.95 | 7.9 | 6.85 | 6.15 | 9.5 | 6.6 | 7.2 | 6.65 | 7 | **5.75** |
| WaveformEW | 22.95 | 22.35 | 22.35 | 21.6 | 35.55 | **20.5** | 22.2 | 22.35 | 27.05 | 22.1 |
| WineEW | 7.55 | 6.3 | 7.35 | 5.6 | 7.8 | **5.4** | 7.45 | 6.6 | 8 | 7.7 |
| Zoo | 8.05 | 7.75 | 7.9 | **5.35** | 9.8 | 5.7 | 7.3 | 5.5 | 6.4 | 5.85 |

**Table 5. Average computational time (in seconds) for different optimizers**

| | ALO-S1 | ALO-S2 | ALO-S3 | ALO-V1 | ALO-V2 | ALO-V3 | bALO1 | bALO2 | GSA | PSO |
|---|---|---|---|---|---|---|---|---|---|---|
| Breastcancer | 1.75 | 1.69 | 1.69 | 1.72 | **1.57** | 1.73 | 1.68 | 1.71 | 3.18 | 3.36 |
| BreastEW | 2.02 | 2.02 | 2.01 | 2.03 | **1.67** | 2.02 | 2.01 | 2.02 | 3.58 | 3.62 |
| Exactly | 2.62 | 2.64 | 2.64 | 2.63 | **2.57** | 2.64 | 2.65 | 2.67 | 5.44 | 4.85 |
| Exactly2 | 3.03 | 2.93 | 2.91 | 2.46 | **2.25** | 2.48 | 2.98 | 2.46 | 5.35 | 4.18 |
| HeartEW | 1.29 | 1.30 | 1.30 | 1.32 | **0.78** | 1.32 | 1.29 | 1.31 | 2.30 | 2.33 |
| Lymphography | 1.19 | 1.21 | 1.20 | 1.24 | **0.70** | 1.22 | 1.20 | 1.23 | 2.15 | 2.13 |
| M-of-n | 2.67 | 2.71 | 2.71 | 2.74 | **2.54** | 2.72 | 2.70 | 2.70 | 5.04 | 4.81 |
| PenglungEW | 3.58 | 3.59 | 3.58 | 3.68 | 3.16 | 3.62 | 3.58 | 3.60 | 2.54 | **2.49** |
| SonarEW | 1.59 | 1.58 | 1.59 | 1.59 | **1.19** | 1.59 | 1.58 | 1.58 | 2.24 | 2.16 |
| SpectEW | 1.34 | 1.33 | 1.34 | 1.36 | **0.87** | 1.36 | 1.33 | 1.36 | 2.27 | 2.29 |
| CongressEW | 1.59 | 1.61 | 1.60 | 1.61 | **0.95** | 1.61 | 1.60 | 1.61 | 2.89 | 2.82 |
| IonosphereEW | 1.56 | 1.56 | 1.55 | 1.56 | **1.26** | 1.55 | 1.56 | 1.56 | 2.53 | 2.46 |
| KrvskpEW | **24.33** | 24.64 | 24.48 | 24.56 | 29.53 | 24.49 | 24.79 | 24.89 | 46.53 | 44.50 |
| Tic-tac-toe | 2.29 | **2.27** | 2.28 | 2.37 | 2.55 | 2.45 | 2.28 | 2.35 | 4.22 | 4.58 |
| Vote | 1.36 | 1.36 | 1.36 | 1.36 | **0.85** | 1.37 | 1.36 | 1.36 | 2.42 | 2.37 |
| WaveformEW | 62.78 | 61.89 | **61.79** | 63.25 | 90.10 | 63.29 | 62.05 | 64.15 | 116.38 | 113.26 |
| WineEW | 1.30 | 1.33 | 1.27 | 1.29 | **0.90** | 1.26 | 1.29 | 1.26 | 2.01 | 1.99 |
| Zoo | 1.30 | 1.30 | 1.30 | 1.31 | **0.79** | 1.30 | 1.29 | 1.28 | 2.07 | 2.11 |

## 5. CONCLUSIONS AND FUTURE WORK

In this paper, we studied the behavior of BALO with six different transfer functions dividing into two classes: s-shaped and v-shaped. The proposed approaches are applied on the feature selection problems. In order to assess the performance of the proposed approaches, 18 well-known UCI benchmark datasets were used, and the results were compared with state-of-the-art approaches. It was shown that the binary ALO approaches equipped with v-shaped transfer functions with their unique method of position update, significantly improve the performance of the original ALO in terms of avoiding local minima and results accuracy.

The simplicity and low-computational cost of the presented algorithms make them suitable to solve a wide range of practical problems such as the those studied in [31 - 34]. For future studies, it is recommended to use the binary algorithm proposed in this wrork for data mining big sensory data sets to reduce the amount of data to be analyzed while building models with efficient interpretability using fewer features.